\PassOptionsToPackage{table}{xcolor}
\documentclass[]{TJURLLAB}

\usepackage{algorithm}
\usepackage{algpseudocode}
\usepackage{wrapfig}
\usepackage{xspace}
\usepackage{tabularx}
\usepackage{array}
\usepackage{nicefrac}
\usepackage{fancyhdr}
\usepackage{fontawesome5}
\usepackage{tablefootnote}
\usetikzlibrary{arrows.meta,calc}
\newcommand{\cW}{\mathcal{W}}           
\newcommand{\cS}{\mathcal{S}}           
\newcommand{\cA}{\mathcal{A}}           
\newcommand{\cE}{\mathcal{E}}           
\newcommand{\cI}{\mathcal{I}}           

\newcommand{\cO}{\mathcal{O}}           
\newcommand{\cC}{\mathcal{C}}           

\newcommand{\cfg}{\chi}                 
\newcommand{\cfgspace}{X}               
\newcommand{\agent}{\mathrm{Ag}}        
\newcommand{\agentspace}{X_{\mathrm{Ag}}}  
\newcommand{\agentst}{\xi}              
\newcommand{\replace}[1]{\chi[#1]}      

\newcommand{\supp}{\operatorname{supp}} 
\newcommand{\Dist}[1]{\Delta\!\left(#1\right)} 

\definecolor{appendixpurple}{RGB}{116,54,176}
\definecolor{appendixlightpurple}{RGB}{248,244,255}

\definecolor{highlightyellow}{RGB}{255,246,205}
\definecolor{highlightframe}{RGB}{232,190,70}
\newcommand{\keysentence}[1]{%
  \begin{center}
    \begin{tcolorbox}[enhanced, width=0.96\linewidth, colback=highlightyellow,
      colframe=highlightframe, boxrule=0.8pt, arc=6pt,
      left=1.2em, right=1.2em, top=0.7em, bottom=0.7em, halign=left]
      #1
    \end{tcolorbox}
  \end{center}
}

\newenvironment{remarkbox}{%
  \begin{tcolorbox}[enhanced, breakable, colback=gray!8, colframe=gray!55,
    boxrule=0.6pt, arc=5pt, left=1.0em, right=1.0em, top=0.5em, bottom=0.5em]
}{%
  \end{tcolorbox}
}

\newenvironment{defbox}{%
  \begin{tcolorbox}[enhanced, breakable, colback=accentcolor!10!white, colframe=accentcolor,
    boxrule=0.8pt, arc=6pt, left=1.2em, right=1.2em, top=0.7em, bottom=0.7em]
}{%
  \end{tcolorbox}
}

\definecolor{takeawaygreen}{RGB}{46,107,94}
\definecolor{takeawaylightgreen}{RGB}{236,246,240}
\newenvironment{takeawaybox}{%
  \begin{tcolorbox}[enhanced, breakable, colback=takeawaylightgreen,
    colframe=takeawaygreen, boxrule=0.7pt, arc=5pt,
    left=1.0em, right=1.0em, top=0.5em, bottom=0.5em]
}{%
  \end{tcolorbox}
}

\newenvironment{exbox}{%
  \begin{tcolorbox}[enhanced, breakable, colback=gray!8, colframe=gray!55,
    boxrule=0.6pt, arc=5pt, left=1.0em, right=1.0em, top=0.5em, bottom=0.5em]
}{%
  \end{tcolorbox}
}

\newenvironment{eqgold}{%
  \begin{tcolorbox}[enhanced, breakable, colback=highlightyellow,
    colframe=highlightframe, boxrule=0.9pt, arc=5pt,
    left=1.2em, right=1.2em, top=0.6em, bottom=0.6em]
}{%
  \end{tcolorbox}
}

\newenvironment{eqpurple}{%
  \begin{tcolorbox}[enhanced, breakable, colback=accentcolor!10!white,
    colframe=accentcolor, boxrule=0.8pt, arc=6pt,
    left=1.2em, right=1.2em, top=0.6em, bottom=0.6em]
}{%
  \end{tcolorbox}
}

\newcommand{\circlednum}[1]{%
  \tikz[baseline={(char.base)}]{%
    \node[shape=circle, draw=accentcolor, fill=accentcolor, text=white,
          font=\footnotesize\bfseries, inner sep=1pt] (char) {#1};}%
}

\newcommand{\circlednumgray}[1]{%
  \tikz[baseline={(char.base)}]{%
    \node[shape=circle, draw=gray!55, fill=gray!28, text=black!80,
          font=\footnotesize\bfseries, inner sep=1pt] (char) {#1};}%
}

\newcommand{\circlednumblack}[1]{%
  \tikz[baseline={(char.base)}]{%
    \node[shape=circle, draw=black, fill=black, text=white,
          font=\footnotesize\bfseries, inner sep=1pt] (char) {#1};}%
}

\makeatletter
\patchcmd{\@thm}
  {\thm@notefont{\fontseries\mddefault\upshape}}
  {\thm@notefont{\bfseries}}
  {\typeout{>>> thm note font patched to bold}}
  {\typeout{!!! thm note font patch FAILED}}
\makeatother

\newcommand{\papercaption}{Generalized Agent Iteration: One Formal Framework for Iterative Policy Improvement and Recursive Self-Improvement}
\fancypagestyle{plain}{\fancyhf{}\fancyfoot[C]{\thepage}}

\title{Generalized Agent Iteration: One Formal Framework for \mbox{Iterative} Policy Improvement and Recursive Self-Improvement}

\author[1]{Hongyao Tang}
\author[1,2]{Yi Ma}
\author[1]{Pengyi Li}
\author[1]{Yifu Yuan}
\affiliation[1]{Tianjin University}
\affiliation[2]{Shanxi University}
\contribution[\faEnvelope]{Contact: \href{mailto:tanghongyao@tju.edu.cn}{tanghongyao@tju.edu.cn}}

\abstract{
\subsubsection*{Abstract}
When we speak of recursive self-improvement (RSI), are we speaking of a
phenomenon, a mechanism, or a prospect? Towards autonomous and evolving
intelligence, RSI is being claimed at many scales, while no single framework
that formally describes these emerging instances exists. Its
counterpart in the classical realm, iterative policy improvement, is
characterized by generalized policy iteration (GPI), a framework of broad
applicability with well-understood theoretical properties, but only where the
update principle and the evaluation base lie outside the agent. In this paper, we propose Generalized Agent
Iteration (GAI), a formal framework that describes iterative policy improvement
and RSI as two cases of a single learning paradigm. GAI
defines the agent as a configuration of modifiable components within a system
and models the learning process as a cycle of \emph{agent evaluation} and
\emph{agent improvement}. Two pivotal dials then distinguish the
instances, in particular the systems that self-identify as RSI in recent work:
whether the improving mechanism is part of the agent and whether the standard
it is measured against is grounded outside it. The former dial
delineates the boundary between GPI and RSI, and the latter determines a
system's polarity as anchored, goal drift, or fully self-referential.
Moreover, we use these coordinates to place existing systems on the same two
axes and make the defects of recursive self-improvement statable one condition
at a time. We see this paper as a first step toward exploring a formal
characterization of RSI that rests on the classical account, makes existing
systems comparable, and provides a principled basis for analyzing and designing
new ones.
}

\begin{document}
\thispagestyle{plain}
\maketitle
\begingroup
\renewcommand{\thefootnote}{}
\makeatletter
\renewcommand{\@makefntext}[1]{\parindent 0pt\noindent\ignorespaces #1}
\makeatother
\footnotetext{Apparently, Generalized Agent Iteration is a salute to the
Generalized Policy Iteration \citep[Section~4.6, pp.~104--105]{sutton2018rl}.}
\endgroup

\section{Introduction}
\label{sec:intro}

Self-improvement is an old yet persistent ambition in artificial
intelligence. It is the premise of Good's ultraintelligent machine, the last
invention that human beings would need to make \citep{good1965}, and the engine
of the later accounts in which a system that improves its own improvement
mechanism compounds into a rapid rise in capability \citep{yudkowsky2013micro}.
How such systems are built, and where they fail, is therefore worth stating
precisely.

Nowadays, self-improvement is claimed at many scales: agents that refine their own outputs
\citep{madaan2023selfrefine,shinn2023reflexion}; systems that rewrite the routine
that improves them, from code-level self-modification
\citep{zelikman2023stop,sica2025} to self-referential agent frameworks
\citep{yin2024goedelagent,zhang2025darwin,kakade2026polaris,zhang2026hyperagents};
research loops that search programs or experiments under fixed evaluators
\citep{novikov2025alphaevolve,lu2024aiscientist}; and, at the far end, systems
that co-evolve the standard they are judged by \citep{iacob2026redqueen} or
propose to do without one \citep{socratic2024}. A recent survey of the area
observes that such labels are used loosely, for ambitions that differ
substantially \citep{chen2026rsisurvey}. Existing works in this space
describe themselves as recursive self-improvement (RSI); we call them
\emph{self-identified RSI}, or \emph{RSI-labeled} for short. What is missing is
\emph{a criterion rather than another example}: a learning-paradigm account that
separates genuine recursive self-improvement from systems that only carry the
label, and says what changes once they do. The classical counterpart of this pursuit is iterative
policy improvement, for which generalized policy iteration (GPI) is the formal
framework: in a finite MDP, its alternation of evaluation and improvement
converges to an optimal policy \citep{sutton2018rl}. GPI assumes that the
improvement mechanism and the reward both lie outside the agent; in the
recursive case they do not, and no account of comparable scope exists
\citep{chen2026rsisurvey,zhang2026rsiblog}.

In this paper, we propose a new formal framework called
\textcolor{accentcolor}{\newterm{Generalized Agent Iteration (GAI)}} with the aim of describing iterative policy improvement and
RSI as two cases of a single learning
paradigm. Beyond these
two cases, it is meant to cover contemporary agent-based learning systems, whose
improving components take the form of code, parameters, or harness, and can be
edited \citep{gao2025survey}. Concretely, we first define the agent within a learning system as a
configuration of modifiable system components.
In analogy with the alternating iteration of GPI, we present the learning
paradigm of GAI as a cyclic iteration of two operations:
\keysentence{%
\begin{itemize}[nosep,leftmargin=1.6em,topsep=2pt,label=$\blacklozenge$]
  \item \emph{Agent evaluation}: multifaceted critics score the agent part
        against an evaluation base.
  \item \emph{Agent improvement}: the modifier proposes a new agent instance
        that the system adopts.
\end{itemize}}

Specific instances of GAI then depend mainly on two open choices, which we call the \emph{two dials}:
\textbf{Dial} \circlednum{1}, whether the mechanism that improves the agent is
part of the agent itself, and \textbf{Dial} \circlednum{2}, whether the standard that
improvement is measured against is grounded outside the agent. The first dial
admits two settings: in GPI the
mechanism that improves the agent stays external, and in recursive
self-improvement it becomes part of the agent. This dial decides which
RSI-labeled systems are RSI in the GAI sense and which are GPI. In GPI the
second dial is fixed as well, with the reward supplied by the world. The setting of the
second dial then determines the \emph{polarity} of self-improvement: anchored,
when the standard remains fixed outside the agent; goal drift, when the agent can
rewrite it; and fully self-referential, when no external standard remains.

Further, we position existing self-improvement systems on the two dials, from
the G\"odel machine and its descendants to co-evolving evaluators and
closed-system proposals. Lastly, we collect the defects of the settings that
leave the anchored end, and state the scope and the open questions that remain.

The main content is summarized below:

\begin{itemize}[leftmargin=*,nosep,topsep=2pt]
  \item We propose a single formal framework that describes both iterative policy
    improvement and recursive self-improvement as two cases of a single learning
    paradigm defined by GAI.
  \item We provide a way of understanding both classical and contemporary
    learning systems through positioning the two dials and agent components
    under GAI.
  \item We present four defects of recursive self-improvement, each tied to a
    condition of classical GPI that a self-improving system violates.
\end{itemize}

The remainder of this paper is organized as follows.
\Cref{sec:background} introduces the background of classical GPI, existing RSI
works, and etc. We deliver the GAI framework in \Cref{sec:gai}, along with
its connection to representative self-improvement systems in
\Cref{sec:connections}. We present the defects of recursive self-improvement in
\Cref{sec:defects}, and close in \Cref{sec:discussion}.

\section{Background}
\label{sec:background}

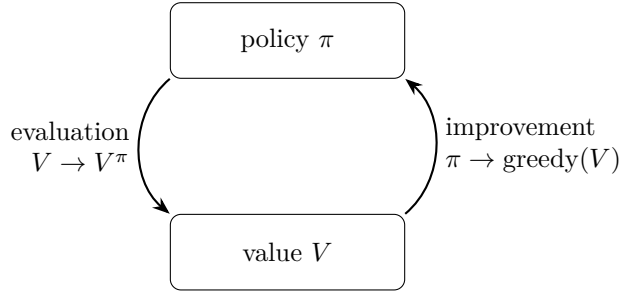
\begin{figure}[t]
  \centering
  \begin{tikzpicture}[>=Stealth,
      box/.style={draw, rounded corners, minimum width=3.1cm, minimum height=1.0cm,
                  align=center}]
    \node[box] (pi) at (0, 1.4) {policy $\pi$};
    \node[box] (val) at (0,-1.4) {value $V$};
    \draw[->, thick] (pi.south west) to[bend right=50]
        node[left, midway, align=right] {evaluation\\ $V\to V^{\pi}$}
        (val.north west);
    \draw[->, thick] (val.north east) to[bend right=50]
        node[right, midway, align=left] {improvement\\ $\pi\to\mathrm{greedy}(V)$}
        (pi.south east);
  \end{tikzpicture}
  \caption{Generalized policy iteration as a cycle: evaluation moves the value
    toward consistency with the current policy ($V\to V^{\pi}$); improvement moves
    the policy toward being greedy in the value
    ($\pi\to\mathrm{greedy}(V)$). Each step may be partial and interleaved
    \citep{sutton2018rl}.}
  \label{fig:gpi}
\end{figure}

\vspace{-0.2cm}\paragraph{Reinforcement Learning and Generalized Policy Iteration (GPI)}
Reinforcement learning \citep{sutton2018rl} is usually formalized as a Markov
decision process (MDP) $(\mathcal{S},\mathcal{A},p,r,\gamma)$: a policy
$\pi\colon\mathcal{S}\to\Delta(\mathcal{A})$
selects actions, and its value is the expected discounted return
$V^{\pi}(s)=\mathbb{E}\big[\sum_{t\ge 0}\gamma^{t}\,r(S_{t},A_{t})\mid S_{0}=s,\;
A_{t}\sim\pi(\cdot\mid S_{t})\big]$, with an action-value $Q^{\pi}(s,a)$
conditioning on the first action. Generalized policy iteration (GPI) is the
shared structure of most single-agent algorithms: \emph{evaluation} moves the
value toward consistency with the current policy,
$V\to V^{\pi}$, and \emph{improvement} moves the policy toward being better in
that value, with $\pi\to\mathrm{greedy}(V)$ the canonical case, as
\Cref{fig:gpi} illustrates.

What makes the framework \emph{generalized} is that neither step is tied to a
concrete implementation: evaluation may be a Bellman backup, a temporal-difference
or Monte-Carlo update, or any operator that drives a value estimate toward
agreement with a policy, and improvement may be greedy but equally softmax,
truncated, or any operator that yields a policy preferred by the current value.
Because only the interaction of an evaluation and an improvement
matters, GPI is a general description of a broad range of iterative
policy-optimization systems, from dynamic programming to model-free reinforcement
learning, and even of such systems when they are not framed as reinforcement
learning at all.


\vspace{-0.2cm}\paragraph{Recursive Self-Improvement Agents}
The literature uses recursive self-improvement for the prospect that a system
improves the very mechanism by which it improves, an idea that dates to Good's
ultraintelligent machine
\citep{good1965} and grounds seed AI \citep{yudkowsky2013micro}, the
instrumental-drive view \citep{omohundro2008basic}, and formal distinctions
between self-modification, self-improvement, and recursive self-improvement
\citep{yampolskiy2015seed,nivel2013bounded}. Its formal formulation is the
G\"odel machine \citep{schmidhuber2003goedeltech,schmidhuber2007goedel}, a
self-referential program that
may rewrite any part of itself whenever it can prove that the rewrite raises a
fixed external utility. Because proof search is intractable, practical
descendants trade proof for empirical validation: the G\"odel Agent
\citep{yin2024goedelagent} pairs a language agent's policy with a self-modifiable
learning algorithm that validates each rewrite on a benchmark; the Darwin
G\"odel Machine \citep{zhang2025darwin} adds archive-based open-ended search;
Polaris \citep{kakade2026polaris} targets small models via auditable policy
repair; and the Red Queen G\"odel Machine \citep{iacob2026redqueen} co-evolves
the evaluator under changing objectives. Even code-level descendants such as
the Self-Improving Coding Agent (SICA) \citep{sica2025} rewrite their own
implementation, yet validate each change against an external coding benchmark
with no gradient update.

A parallel, largely empirical line refines today's language agents in closed
loops without retraining: Self-Refine \citep{madaan2023selfrefine} and Reflexion
\citep{shinn2023reflexion} improve an agent's own outputs, and coding agents such
as STOP \citep{zelikman2023stop} modify their own implementation. At the scale
of a research loop, AlphaEvolve searches programs under fixed evaluators
\citep{novikov2025alphaevolve} and the AI Scientist runs idea-to-paper cycles
\citep{lu2024aiscientist}. One survey organizes such systems by what
evolves, be it parameters, prompts, memory, tools, or scaffolds
\citep{gao2025survey}, and another separates bounded self-refinement, which is
convergent and evaluable, from open-ended recursive self-improvement, ordering
candidate verification signals from formal verifiers down to intrinsic
self-assessment \citep{chen2026rsisurvey}. As deployed, these RSI-labeled
loops, and the scaffold self-modifiers above, keep some component fixed outside the loop,
be it model weights, a prompt template, an outer loop, or a verifier; the agent
then improves within, but not beyond, that fixed mechanism, so the attainable
quality is bounded by it. Most such loops are
therefore instances of bounded self-improvement inside a fixed mechanism.
What the literature calls strong RSI is, under GAI, the case in which the
improver is an agent component; that is the first dial \citep{zhang2026rsiblog}.

\vspace{-0.2cm}\paragraph{The Absence of a Unified Formal Framework}
A recent roadmap survey grades self-improvement systems by how much of the
improvement responsibility the AI internalizes, from executing human-specified
updates to revising the mechanism that governs later improvement
\citep{duan2026lastai}. Taken together, these strands still lack a unified formal account in which GPI
is a special case, recursive self-improvement is a nearby and definable case, and
the guarantees of the former can be located, one by one, as they fail when the
underlying scheme is made self-referential. Existing terminologies, such as
G\"odel agents, meta-learning, and agentic loops, do not make explicit that
these processes live on one continuum. We develop such an account in the next section.

The classical methods and the self-improvement systems above differ along two
informal axes that we formalize in the next section: whether the mechanism that
improves the agent lies inside the agent, and whether the standard that
improvement is measured against stays grounded. \Cref{tab:preview} places a few
representative methods on these axes as a preview.

\begin{table}[t]
  \centering
  \small
  \setlength{\tabcolsep}{5pt}
  \caption{A schematic preview of classical policy-iteration methods and of
    self-improvement agents along three axes, two of which the next section
    formalizes as dials:
    what the method modifies, whether its improvement mechanism lies inside the
    agent, and whether its evaluation basis is fixed outside it.}
  \label{tab:preview}
  \vspace{-0.2cm}
  \begin{tabularx}{\textwidth}{@{} >{\raggedright\arraybackslash}p{2.5cm}
                     | >{\raggedright\arraybackslash\hsize=1.0\hsize}X
                     >{\raggedright\arraybackslash\hsize=1.1\hsize}X
                     >{\raggedright\arraybackslash\hsize=1.0\hsize}X
                     | >{\raggedright\arraybackslash\hsize=0.9\hsize}X @{}}
    \toprule
    \textbf{Method}
      & \textbf{\circlednumgray{0}~Modification}
      & \textbf{\circlednumgray{1}~Improvement Mechanism}
      & \textbf{\circlednumgray{2}~Evaluation Basis}
      & \textbf{Guarantee} \\
    \midrule
    Policy / value iteration \citep{sutton2018rl}
      & \begin{itemize}[nosep,leftmargin=*,topsep=0pt]
          \item policy
          \item value
        \end{itemize}
      & external, fixed (greedy; Bellman backup)
      & environment reward
      & converges to an optimal policy \\
    \midrule
    DQN \citep{mnih2013dqn}
      & \begin{itemize}[nosep,leftmargin=*,topsep=0pt]
          \item deep Q-network (action values)
          \item implicit greedy policy
        \end{itemize}
      & external, fixed (temporal-difference update; greedy)
      & environment reward
      & no exact guarantee under function approximation \\
    \midrule
    Reflexion / Self-Refine \citep{shinn2023reflexion,madaan2023selfrefine}
      & \begin{itemize}[nosep,leftmargin=*,topsep=0pt]
          \item own outputs
          \item reflection memory
        \end{itemize}
      & external, fixed (retry loop; the model rewrites its own output)
      & task feedback / self-critique
      & empirical gains, bounded by the fixed loop \\
    \midrule
    STOP \citep{zelikman2023stop}
      & \begin{itemize}[nosep,leftmargin=*,topsep=0pt]
          \item own optimizer (for code generation)
        \end{itemize}
      & internal (the optimizer rewrites itself)
      & validation examples
      & empirical gains \\
    \midrule
    G\"odel Agent / Darwin G\"odel Machine
      \citep{yin2024goedelagent,zhang2025darwin}
      & \begin{itemize}[nosep,leftmargin=*,topsep=0pt]
          \item own code
          \item policy and improvement mechanism
          \item open-ended archive search (Darwin)
        \end{itemize}
      & internal (the self-modification routine is modifiable)
      & external benchmark
      & each rewrite empirically validated \\
    \midrule
    Red Queen G\"odel Machine \citep{iacob2026redqueen}
      & \begin{itemize}[nosep,leftmargin=*,topsep=0pt]
          \item the agent
          \item its evaluator
        \end{itemize}
      & internal (the agent and its evaluator co-evolve)
      & co-evolving evaluator; no fixed external basis
      & no guarantee relative to a fixed external basis \\
    \bottomrule
  \end{tabularx}
  \vspace{-0.2cm}
\end{table}

\section{The Generalized Agent Iteration (GAI) Framework}
\label{sec:gai}

Although GPI describes a broad class of policy-improvement methods, it fails to describe
the more complex systems that improve recursively, in which the mechanism
that improves the agent is itself part of what the agent can change. To this
end, we propose a formal framework, which we call Generalized Agent Iteration
(GAI), a learning paradigm that extends GPI to describe agent-based
learning systems whose improving components can themselves be edited. GAI keeps the alternating cycle of evaluation and improvement at its center,
and treats GPI and recursive self-improvement as two instances of it. The two instances differ in two choices, which we call the
\emph{two dials} of the framework: \circlednum{1} whether the mechanism that
improves the agent lies inside the agent, and \circlednum{2} whether the
standard that improvement is measured against stays grounded in what lies
outside it.

The rest of this section turns the two dials into formal definitions: it
defines the system and the agent (\Cref{sec:vocab}) and reads GPI and
recursive self-improvement as two settings of the dials
(\Cref{sec:instances}). It closes by characterizing the polarity of
self-improvement and reducing recursive self-improvement back to GPI
(\Cref{sec:dials}).

\subsection{The Self-Improving System and Agent}
\label{sec:vocab}

A self-improving system acts in a world and is measured against a goal. We
keep the world as a Markov decision process $\cW=(\cS,\cA,p,r,\gamma)$, and write
$G$ for an external goal that the system is expected to serve. Both belong
to the environment, and neither is part of the system itself.

The objective of the system is to serve the external goal $G$, whose scalar
instance in the world is the reward $r$. The configuration that serves it best is
\begin{equation}
  \chi^{\star}\in\arg\max_{\chi\in\cfgspace}\;
  \mathbb{E}\Big[\textstyle\sum_{t\ge0}\gamma^{t}r(S_{t},A_{t})\;\Big|\;\chi\Big],
  \label{eq:objective}
\end{equation}
This objective is fixed by the world and the goal alone, not by the system
itself. More generally, the world and the goal may change over time on their
own; we hold both fixed in what follows, since such changes originate outside
the system and leave the analysis below unaffected.
\begin{remarkbox}
\textbf{Remark 1 (goal $G$ in the world v.s., self-proposed tasks).} Some self-improving systems have the agent
propose its own tasks, questions, or goals as a means of learning. We treat this
as a particular form of agent improvement, since such self-set tasks remain
subordinate to an external goal that is objective and cannot be edited by the
agent.
\end{remarkbox}

Let $\chi$ be a system made of components, with $\cO=\{\pi,V,m,U,\rho\}$ the
set of components and $\cC_o$ the content space of component $o$; the system
then lies in the space $\chi \in \cfgspace=\prod_{o\in\cO}\cC_o$.
Let $\agent\subseteq\cO$ be the set of components that form the agent; an
agent instance is an assignment $\agentst\in\agentspace=\prod_{o\in\agent}\cC_o$
of contents to these components. Moreover, we write $\replace{\agentst}$ for
the system obtained by replacing the agent part of $\cfg$ with $\agentst$.
We use $\Dist{Y}$ to denote the set of probability distributions over a space $Y$.
We define the system and its agent as follows.

\begin{defbox}
\textbf{Definition 1 (System and Agent).} Over an external world
$\cW=(\cS,\cA,p,r,\gamma)$ and a goal $G$, a system is a configuration
$\cfg=(\pi,V,m,U,\rho)$, a tuple of the current contents of five
components:
\begin{itemize}[nosep,leftmargin=1.2em,topsep=0pt]
  \item $\pi:\cS\to\Dist{\cA}$: the \newterm{policy} that chooses world actions;
  \item $V:\cS\to\R$: the \newterm{action critic} that evaluates how well $\pi$ acts;
  \item $m:\cfgspace\to\Dist{\agentspace}$: the \newterm{modifier} that proposes how the system should change;
  \item $U:\cfgspace\to\R$: the \newterm{modification critic} that evaluates whether a proposed change serves $G$;
  \item $\rho$: the \newterm{evaluation base}, the standard that the critics measure against.
\end{itemize}
All five roles belong to a system; an instance may leave some of them unused,
as the modification critic $U$ is when the modifier is fixed. The agent is
the modifiable parts inside the system. Its membership varies across systems, such as $\agent=\{\pi,V\}$ when the modifier is
fixed, or $\agent=\{\pi,V,m,U,\rho\}$ when the system may also rewrite its
modifier and even the evaluation standard.
\end{defbox}

\begin{remarkbox}
\textbf{Remark 2 (roles and substrates).} The five components are functional
roles, not objects: $\pi$ acts, $V$ and $U$ score, $m$ produces the changes,
and $\rho$ is the standard. Systems edit substrates, such as prompts, memory,
tools, weights, or code; a substrate is one realization of a role, read
through the role whose content it changes. A policy may be realized entirely
as code, with no model at deployment \citep{hu2026coap}.
\begin{center}
\footnotesize
\begin{tabularx}{\linewidth}{@{} >{\raggedright\arraybackslash}X >{\raggedright\arraybackslash}X @{}}
\toprule
\textbf{Substrate} & \textbf{Read as} \\
\midrule
weights; prompts, memory, tools, skills & $\pi$: how the agent acts \\
training data, curricula, collected experience & material for the next $\pi$ \\
trainer, optimizer, scheduler, task proposer & $m$: what produces the changes \\
evaluator, test sets, release rules & the standard ($V$, $U$, $\rho$) \\
sandbox, schemas, type contracts & constraints on $m$ \\
an artifact that no role reads or writes & the world, outside the system \\
\bottomrule
\end{tabularx}
\end{center}
One role may span several substrates, and one substrate may carry several
roles: a memory store holds a tool description in one entry, an acceptance
criterion in another. The dials read roles rather than substrates, so three
surfaces \citep{tan2026metarsi} or nine \citep{duan2026lastai} are read the
same way, and the tuple may be extended with further components.
\end{remarkbox}

\begin{remarkbox}
\textbf{Remark 3 (the interior of the modifier).} The modifier is a
functional role, and its content may be a pipeline: a proposer of tasks or
questions to learn from, a store of the resulting experience, and a trainer
that turns the experience into the next agent instance, together with a
schedule over these parts. A self-training loop, in which the system sets its
own tasks, collects the experience, and learns from it, is read as a modifier
of this kind, and whether that modifier is itself an agent component is the
question of the first dial below, not of its internal organization. Grading
the interior, for instance by how many of these decisions the system makes,
is the work of the level-based accounts discussed in \Cref{sec:connections},
and it leaves both dials unaffected. Self-proposed tasks move the second dial
only through the standard that scores the changes they produce.
\end{remarkbox}

Definition 1 fixes the content of a system but not how it evolves. We next
define the process by which a system changes over time.

\begin{defbox}
\textbf{Definition 2 (Generalized Agent Iteration).} A generalized agent
iteration is a process that alternates two operations on the agent part of a
system:
\begin{itemize}[nosep,leftmargin=1.2em,topsep=0pt]
  \item \emph{Agent evaluation}: the critic in use scores the agent part
        against the evaluation base $\rho$, with $V$ at the policy level and
        $U$ at the agent level;
  \item \emph{Agent improvement}: the modifier produces a new agent instance,
        $\agentst\sim m(\cdot\mid\cfg)$, which the system adopts,
        $\cfg\leftarrow\cfg[\agentst]$.
\end{itemize}
\end{defbox}

\begin{figure}[t]
  \centering
  \begin{tikzpicture}[>=Stealth,
      box/.style={draw, rounded corners, minimum width=2.0cm, minimum height=0.8cm,
                  align=center}]
    \node[box] (pi)    at ( 0.0,  2.4) {policy $\pi$};
    \node[box] (crit)  at (-2.0, -1.0) {critic $V/U$};
    \node[box, dash dot] (m) at ( 2.0, -1.0) {modifier $m$};
    \node[font=\small] (rho) at (-4.5, -1.0) {(evaluation base $\rho$)};
    \node[font=\small, gray!80, align=center] (env) at (-4.5, -2.25)
        {environment\\ and goal};
    \draw[->, thick] (env) -- (rho);
    \draw[->, thick] ($(pi.south west)+(0,0.2)$) to[out=195, in=105, looseness=1.0]
        node[left, midway, font=\small] {evaluation}
        ($(crit.north)+(-0.4,0)$);
    \draw[->, thick] (crit.south) to[bend right=55]
        node[below, midway, font=\small] {feedback} (m.south);
    \draw[->, thick, dashed, gray!60] (m.north) to[bend right=30]
        node[above, midway, font=\small] {evaluation} (crit.north);
    \draw[->, thick, dashed, gray!60] (m.west) --
        node[below, midway, font=\small] {improvement} (crit.east);
    \draw[->, thick] ($(m.north)+(0.4,0)$) to[out=75, in=-15, looseness=1.0]
        node[right, midway, font=\small] {improvement}
        ($(pi.east)+(0,-0.2)$);
    \draw[->, thick, dashed, gray!60] (m) to[loop right, looseness=6]
        node[right, font=\small, align=left] {improvement:\\ modifier may\\ rewrite itself} (m);
  \end{tikzpicture}
  \caption{Generalized agent iteration as a cycle of policy, critic, and
    modifier. The critic evaluates the policy and the
    modifier against the base $\rho$ and returns feedback; the modifier then
    produces improvements, including possible rewrites of the critics and
    itself (gray dashed, conditional on the setting; the dash-dot border of
    the modifier marks that it belongs to the agent only when $m\in\agent$).
    With the modifier fixed outside the agent the cycle reduces to GPI
    (\Cref{fig:gpi}); with the modifier inside the agent it is recursive
    self-improvement.}
  \label{fig:gai}
\end{figure}
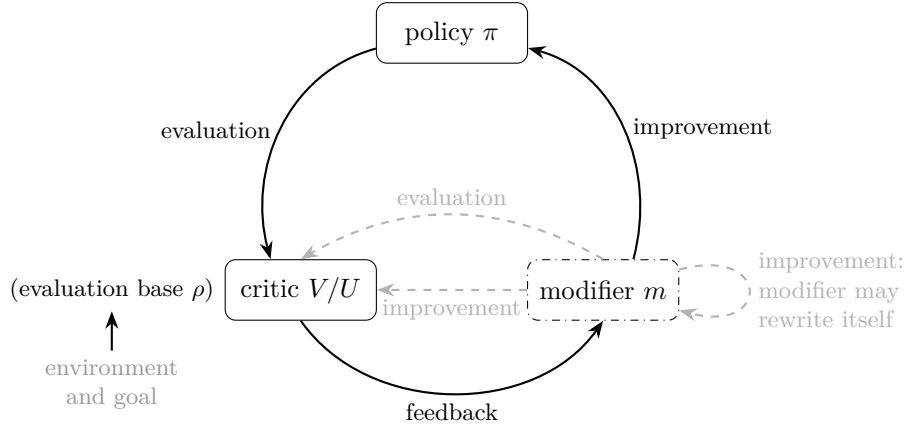

Definition 1 leaves two choices open: whether the mechanism that improves the
agent is among these components, and what the evaluation base points at. These are the two dials of the
framework, and each formalizes one way in which GPI and recursive
self-improvement differ.

\circlednum{1}~\textbf{First dial: whether the mechanism that improves the
agent lies inside the agent.} The agent changes only through the modifier:
at each step $m$ draws a new agent instance $\agentst\sim m(\cdot\mid\cfg)$,
and the system moves to $\replace{\agentst}$.
\begin{itemize}
  \item If $m\notin\agent$, its content is fixed and $m$ acts as
an external improvement mechanism, the abstract evaluation-and-improvement
operator that GPI iterates. 
  \item If $m\in\agent$, the system may rewrite its own modifier, so recursion closes at $m$ with no external meta-layer.
\end{itemize}

\circlednum{2}~\textbf{Second dial: whether the standard that improvement is
measured against stays grounded.} Each evaluation regresses toward a standard,
its evaluation base; we treat this base as a single standard $\rho$ and write
its instances $\rho_V$ and $\rho_U$ only where the two critics must be told
apart. The base is a component of the system, and what the second dial asks
about is where its content comes from and whether the agent may rewrite it. An
evaluation is \textit{grounded} when its base draws its content from outside
the system, the world and the goal, and when the base itself lies outside the
agent. The agent then cannot rewrite the standard it is measured against. A base that draws its content from outside but that
the agent may rewrite can move under the system's own edits; a base with no
external content at all constrains the loop by nothing beyond self-consistency. In a system that
improves recursively, what matters is whether the base that its
self-improvement is measured against stays grounded, which we examine in later
subsections.

\begin{exbox}
\textbf{Running Example (a learning coding agent).} Consider an agent that
writes code, whose configuration is $\cfg=(\pi,V,m,U,\rho)$: the policy
$\pi$ generates candidate solutions, the action critic $V$ scores them
against the current tests, the modifier $m$ proposes changes to the agent,
such as rewriting the prompt that drives the policy or the routine by which
the agent revises its own solutions, and the modification critic $U$ evaluates
each proposed change. We keep this agent as the
concrete referent for the roles above, and read off its two dial settings below.
\end{exbox}

\subsection{GPI and Recursive Self-Improvement as Two GAI Instances (Dial 1)}
\label{sec:instances}

We now re-introduce GPI and recursive self-improvement as two instances of
the system definition in the previous subsection. They differ only in how the two dials
are set: for GPI the modifier lies outside the agent and the evaluation base
is grounded in the world reward; for recursive self-improvement the modifier
is itself an agent component. Everything else in the description is shared.

\textbf{GPI as the anchored instance.} The agent consists of the policy and
the action critic, $\agent=\{\pi,V\}$, and the modifier is not part of the
agent, $m\notin\agent$. The content of $m$ is fixed to an evaluation operator
and an improvement operator, written $\cE_m$ and $\cI_m$ to mark that the
update principle is the content of $m$; since $m$ is fixed, both operators are
fixed as well. Through $m$ the agent changes componentwise as
\begin{equation}
  V\;\leftarrow\;\cE_m(V,\pi;\rho_V)
  \qquad\text{(evaluation)},\qquad
  \pi\;\leftarrow\;\cI_m(V)
  \qquad\text{(improvement)},
  \label{eq:gpiupdate}
\end{equation}
while $m$ itself is unchanged. The only evaluation in use is the action
critic; the modification critic $U$ is present but not exercised, since $m$ is
fixed and no alternative modification is scored. The base $\rho_V$ of the
action critic is grounded, with content the world reward $r$. When $\cE_m$ is
policy evaluation and $\cI_m$ is greedy, the two updates form exactly policy
or value iteration, which in a finite MDP converges to an optimal policy
$\pi^{\star}$ under standard assumptions. The objective of
\Cref{eq:objective} is then maximized by that same policy, with the action
critic at its value $V^{\pi^{\star}}$.

\textbf{Recursive self-improvement as the self-modifying instance.} Now the
modifier is part of the agent, $m\in\agent$, canonically
$\agent=\{\pi,V,m,U\}$. This is the GAI sense of RSI: the location of the
improver, not the object it improves, is the criterion. The step is unchanged
and now reads
\begin{eqgold}
\begin{equation}
\begin{aligned}
&\text{policy update:}\quad
  V\;\leftarrow\;\cE_m(V,\pi;\rho_V),\quad
  \pi\;\leftarrow\;\cI_m(V)\\[2pt]
&\text{agent update:}\quad
  \agentst\sim m(\cdot\mid\cfg),\quad
  \cfg\leftarrow\replace{\agentst},\quad
  m'\leftarrow(\agentst)_m,\quad
  U'\leftarrow(\agentst)_U
\end{aligned}
\label{eq:rsidynamics}
\end{equation}
\end{eqgold}
The operators in the first line are subscripted by $m$ to mark that they are
decided by the current modifier rather than fixed externally: the
policy-level updates of GPI become, in RSI, content that $m$ may realize.
The second line is the recursion itself: $(\agentst)_m$ and $(\agentst)_U$
are the modifier and modification-critic slots of the proposal, so the next
modifier and the next critic are produced by the current one, recursion
closes at $m$, and no external meta-layer updates it. Every agent
component changes only through $m$, and $m$ may run the GPI-style alternation
above, but it may equally rewrite itself, replace its critic, or change the
rule by which it chooses. Our definition leaves the second dial, whether the base
that governs this recursion stays grounded, open; different self-improvement
systems adopt different choices, and we examine the consequences next.

\textbf{The reference ideal of a self-improver.} Nothing in this
description guarantees that $m$ improves toward anything external; improving
is a property of $m$'s content, not of the framework. If $m$ improves toward
its base, its content $(m,U)$ must satisfy a pair of
self-consistency conditions, which we state as a reference ideal.

Write $\rho_U(\cfg,\agentst)$ for the value that the base assigns to adopting
agent instance $\agentst$ from $\cfg$. Then
\begin{eqpurple}
\begin{equation}
\begin{aligned}
&\text{agent evaluation:}\quad
  U(\cfg)=\mathbb{E}_{\agentst\sim m(\cdot\mid\cfg)}
  \Big[\rho_U(\cfg,\agentst)+\gamma\,U\big(\replace{\agentst}\big)\Big]\\[2pt]
&\text{agent improvement:}\quad
  \supp m(\cdot\mid\cfg)\subseteq
  \arg\max_{\agentst\in\agentspace}
  \Big[\rho_U(\cfg,\agentst)+\gamma\,U\big(\replace{\agentst}\big)\Big]
\end{aligned}
\label{eq:selfconsistent}
\end{equation}
\end{eqpurple}
where the first condition calibrates the modification critic to the base and
the second requires the modifier to propose only agent instances that improve
in the critic's estimate. These are characterization conditions, not enforced
updates: they describe the internal structure a self-improver must have,
and no external mechanism guarantees them. Whether the base that appears here
is grounded is the polarity question we turn to next.

\begin{exbox}
\textbf{Running Example (Anchored Learning v.s., Recursive Self-Improvement).}
The coding agent above realizes the two instances just described.
\begin{itemize}[leftmargin=*,nosep]
  \item \textbf{Anchored learning (GPI style).} The agent is
    $\agent=\{\pi,V\}$, and $m$ is a fixed external update routine, for
    instance a PPO-style loop \citep{schulman2017ppo} that trains the policy
    against the reward of passing its unit tests. The system improves within
    the routine but cannot change the routine itself.
  \item \textbf{Recursive self-improvement.} The agent is
    $\agent=\{\pi,V,m,U\}$, and the proposal $\agentst$ that $m$ draws
    carries the next modifier, so the system may rewrite the very routine
    that improves it. STOP is one concrete instance, in which the improvement
    routine rewrites its own code against a fixed set of validation examples
    \citep{zelikman2023stop}.
\end{itemize}
\end{exbox}

\subsection{Polarity of Self-Improvement: Anchored, Goal Drift, and Fully Self-Referential (Dial 2)}
\label{sec:dials}

The second dial applies to the base of the action critic and to the base of
the modification critic alike. It separates polarities only when the modifier
is part of the agent: in GPI the modification critic is never exercised, so the
only base in use is that of the action critic, grounded in the world reward,
and the polarity entry for GPI is Anchored. For $m\in\agent$, the second dial
decides the \newterm{polarity} of self-improvement, namely whether the
evaluation base $\rho$ stays grounded; \Cref{tab:polarity} lists the three
states it can take.

\begin{table}[t]
  \centering
  \small
  \caption{Polarity of self-improvement for systems with the modifier in the
    agent, determined by the state of the evaluation base $\rho$; the last
    column lists representative systems.}
  \label{tab:polarity}
  \vspace{-0.2cm}
  \begin{tabularx}{\textwidth}{@{} >{\raggedright\arraybackslash}p{2.5cm}
                     | >{\raggedright\arraybackslash}p{3.1cm}
                     >{\raggedright\arraybackslash\hsize=1.0\hsize}X
                     >{\raggedright\arraybackslash\hsize=1.1\hsize}X @{}}
    \toprule
    \textbf{Polarity}
      & \textbf{State of $\rho$}
      & \textbf{Consequence}
      & \textbf{Representative systems} \\
    \midrule
    Anchored
      & external content, $\rho\notin\agent$, so the agent cannot rewrite it
      & Improves toward a fixed external standard; a modifier satisfying the
        self-consistency conditions above tracks it.
      & G\"odel machine \citep{schmidhuber2007goedel}; STOP
        \citep{zelikman2023stop}; MetaRSI \citep{tan2026metarsi} \\
    \midrule
    Goal Drift
      & external content, but $\rho\in\agent$, so the agent may rewrite it
      & The standard itself can be rewritten, so self-improvement can drift
        away from the external goal.
      & Red Queen G\"odel Machine \citep{iacob2026redqueen} \\
    \midrule
    Fully Self-Referential
      & no external content: empty, or $\rho$ depends on $U$
      & No external signal reaches the self-improvement loop, and the critic
        equation reduces to internal self-consistency.
      & Socratic learning \citep{socratic2024} (position paper) \\
    \bottomrule
  \end{tabularx}
  \vspace{-0.2cm}
\end{table}

\begin{remarkbox}
\textbf{Remark 4 (learned proxies).} An external judge or reward model is
anchored, whatever data it was trained on: the dial asks where the standard
sits, not how faithful it is to the goal. A proxy can be anchored and still be
satisfied without serving the goal; the same judge as an agent component is the
Goal Drift case.
\end{remarkbox}

The rows of \Cref{tab:polarity} are ordered by how much anchoring remains. The
middle row is named \emph{Goal Drift} for the drift of the system away from its
fixed goal, not for a change in the goal itself: what moves is the standard
that measures progress toward it. In current practice, most demonstrated
RSI systems fall into the Anchored row; the two lower rows are so
far represented mainly by boundary cases and by position papers.

The two dials, taken together, place every instance. Setting the modifier
outside the agent and fixing the only base in use to the world reward gives the
GPI loop with $\agent=\{\pi,V\}$, of which the PPO-trained coding agent in
the running example is an instance. Keeping the modifier inside the agent
while keeping $\rho$ grounded gives anchored recursive self-improvement, as
in STOP, with its fixed validation examples, or the G\"odel machine, whose
proof gate restricts which rewrites are tried. Letting $\rho$ enter the
agent, or vanish or point at itself, gives the unanchored end, whose
consequences we examine after mapping existing systems onto the dials.
Freezing the modifier as an external procedure and removing the modification
critic and its base $\rho_U$ recover the classical loop over policy and value.

\section{Connections to Existing Self-Improvement Systems}
\label{sec:connections}

The two dials of the previous section give the main families discussed
in \Cref{sec:background}\footnote{Output-level refinement loops such as
Reflexion and Self-Refine are outside Table 3: the optimized object is the
output, not an agent component, and the base is the model's own critique.},
and the recent systems that extend them, a uniform
reading. The reading is the GAI classification, not a report of the
literature's labels: several RSI-labeled systems keep the modifier outside the
agent and fall into the fixed-outer-loop row. We state that reading here for the
fixed-loop practice that dominates current work, for the closest formal
counterpart of recursive self-improvement, and for the empirical families that
follow it; \Cref{tab:reading} summarizes it.

\begin{table}[t]
  \centering
  \footnotesize
  \setlength{\tabcolsep}{4pt}
  \caption{Existing systems under the GAI reading: the agent set, whether the
    modifier is modifiable (Dial 1), the resulting polarity and evaluation base (Dial 2),
    and the classification. The fixed-outer-loop row includes RSI-labeled systems;
    under GAI their modifier stays outside the agent. The evaluator of the Red Queen
    row is the modification critic $U$, which is an agent component whenever it is modifiable.}
  \label{tab:reading}
  \vspace{-0.2cm}
  \begin{tabularx}{\textwidth}{@{} >{\raggedright\arraybackslash}p{2.5cm}
                     | >{\raggedright\arraybackslash}p{2.6cm}
                     >{\raggedright\arraybackslash\hsize=1.0\hsize}X
                     >{\raggedright\arraybackslash\hsize=1.0\hsize}X
                     | >{\raggedright\arraybackslash}p{2.5cm} @{}}
    \toprule
    {\centering\textbf{System}\par}
      & {\centering\textbf{Agent $\agent$}\par}
      & {\centering\footnotesize\textbf{\shortstack{Dial~\circlednum{1}\\[1pt] $m$ modifiable?}}\par}
      & {\centering\footnotesize\textbf{\shortstack{Dial~\circlednum{2}\\[1pt] Polarity}}\par}
      & {\centering\textbf{Classification}\par} \\
    \midrule
    Policy/value iteration, PPO \citep{sutton2018rl,schulman2017ppo}
      & $\{\pi,V\}$
      & no; $m$ is the fixed update principle (Bellman backup and greedy improvement; the clipped policy update for PPO)
      & Anchored; the evaluation base is the environment reward
      & (Anchored) GPI \\
    \midrule
    Fixed outer loops \citep{gao2025survey,zheng2026dreamrsi,neohorse2026,liu2026solpi,hu2026coap,lang2026embodirsi,yang2026terminalrsi,tang2026trinity}
      & the optimized object ($\pi$ or a harness)
      & no; the outer loop is fixed by its designers
      & Anchored; the evaluation base is the objective or validation metric set outside
      & (Anchored) GPI \\
    \midrule
    G\"odel machine \citep{schmidhuber2003goedeltech,schmidhuber2007goedel}
      & $\{\pi,m\}$ (solver and its rewriter)
      & yes; $m$ is the rewriting routine that edits the program when a proof succeeds
      & Anchored; the evaluation base is a fixed external utility, with a proof gate on rewrites
      & Anchored RSI (guarded) \\
    \midrule
    G\"odel Agent, Darwin G\"odel Machine, STOP, SICA, Polaris, Hyperagents, AIDE$^{2}$ \citep{yin2024goedelagent,zhang2025darwin,zelikman2023stop,sica2025,kakade2026polaris,zhang2026hyperagents,srikanth2026aide2}
      & $\{\pi,m\}$ (the harness; model weights and benchmark outside)
      & yes; $m$ is the self-modification routine that edits prompts, memory, tools, or code
      & Anchored; the evaluation base is an external benchmark or validation set
      & Anchored RSI \\
    \midrule
    MetaRSI \citep{tan2026metarsi}
      & $\{\pi,m\}$ (data, weights, and scaffold, and the schedule over them)
      & yes; the operators' proposal policies and their schedule are revised at a meta level
      & Anchored; a sealed evaluator, task set, and release gate stay outside every write surface at every level
      & Anchored RSI (a fixed meta level ends the hierarchy) \\
    \midrule
    Red Queen G\"odel Machine \citep{iacob2026redqueen}
      & $\{\pi,m,U\}$ (the agent and its evaluator, which co-evolve)
      & yes; $m$ rewrites both the agent and its evaluator
      & Goal Drift; the evaluation base is updated during the process, so no fixed external standard remains
      & Drifting RSI \\
    \midrule
    Closed-system proposals \citep{socratic2024}
      & all components, $\rho$ included
      & yes; $m$ may rewrite any component, the base included
      & Fully Self-Referential; the evaluation base is empty or self-referential, so no external signal remains
      & Fully Self-Referential RSI \\
    \bottomrule
  \end{tabularx}
  \vspace{-0.2cm}
\end{table}

\vspace{-0.2cm}\paragraph{\protect\circlednumblack{1}~Fixed Outer Loops}
A large body of practice improves a model or a harness inside a pipeline whose
control loop is fixed by its designers: proposal, update, and selection are
scripted, and the system cannot change the loop itself \citep{gao2025survey}.
Research loops such as AlphaEvolve and the AI Scientist are the same setting:
the search is scripted and the evaluator is fixed
\citep{novikov2025alphaevolve,lu2024aiscientist}. This is the Anchored
setting with the modifier frozen outside the agent: the mechanism of
improvement is not itself changed. The first row of \Cref{tab:reading} is the
special case in which the agent holds the policy and its action critic; in the
pipelines above, the agent is typically a single optimized object, a model or a
harness. Recent systems fill out the setting and sharpen a distinction that
the literature blurs. Dream-RSI improves the policy that steers its search
while the revision loop, the underlying agent, and the evaluator stay fixed,
and its guarantee that the revised policy is never worse on the replay score
is monotonicity of the classical kind, available exactly because the modifier
and the base both sit outside the agent \citep{zheng2026dreamrsi}. The
literature calls this meta-layer self-improvement, since the object it improves
governs later improvements. GAI has no separate meta-layer setting: the first
dial asks who produces the changes, so a system that improves a meta-level
object remains a fixed outer loop as long as the producer stays fixed.
NeoHorse-1 runs one pass of a fixed post-training pipeline on traces of its own
deployment, measured on a held-out suite \citep{neohorse2026}. SoL-Pi scales
this loop over 535 environments and freezes candidates before a held-out
evaluation that never flows back \citep{liu2026solpi}. Terminal-RSI fixes its
five-stage schedule \citep{yang2026terminalrsi}, and The Trinity keeps its
Creator and Evolver outside the evaluated harness under a fixed topology
\citep{tang2026trinity}; under GAI both are fixed outer loops, not RSI in the
GAI sense.
The same reading covers the recent
embodied systems: a policy written entirely as code by an offline coding
agent \citep{hu2026coap} and a weight-level adaptation driven by simulated
rollouts under fixed round and data quotas \citep{lang2026embodirsi} keep both
the improver and the base outside the agent, and join this family rather than
adding a setting.

\vspace{-0.2cm}\paragraph{\protect\circlednumblack{2}~The G\"odel Machine and Its Descendants}
The G\"odel machine \citep{schmidhuber2003goedeltech,schmidhuber2007goedel} is
the closest formal neighbor: a program that may rewrite any part of itself when
it proves that the rewrite raises a fixed external utility. In the present
terms it is an anchored self-improver: the modifier lies in the agent, the base
is the external utility, and a proof gate restricts which rewrites are
admissible. The G\"odel-agent family keeps the pattern while replacing proof
with empirical validation: these systems rewrite their own code or harness and
are judged on a benchmark that stays outside the agent, so they also sit in the
Anchored row \citep{yin2024goedelagent,zhang2025darwin,kakade2026polaris,zelikman2023stop,sica2025,srikanth2026aide2}.
Memento 3 freezes the underlying model and accepts a rulebook or world-model
revision only when replay reproduces the observations \citep{zhao2026memento3}.
Hyperagents goes further: it places the task agent and the meta agent that
improves it in a single editable program, so that the modification procedure
itself can be rewritten, while the evaluation remains an external benchmark
\citep{zhang2026hyperagents}. Since only the scaffold is modifiable, such
systems are bounded in what they can improve. MetaRSI widens the editable set
to data, weights, and scaffold, and revises the schedule by which its
operators are composed, while a sealed evaluator, the task set, and the
release gate stay outside every write surface at every level, so the anchoring
is enforced by construction rather than assumed \citep{tan2026metarsi}. Its
meta level, which revises the schedule, is in turn fixed, so the hierarchy
terminates at a level the system does not rewrite. In the same months, a wave
of harness- and weight-level work keeps the scoring outside the agent
while spanning both settings of the first dial
\citep{ye2026harnesszero,dong2026mosib,tian2026physevo,zhao2026rsi4eo};
RSIGym provides such an environment \citep{meng2026rsigym}.

\begin{remarkbox}
\textbf{Remark 5 (language agents as model and harness).} Contemporary language
agents are described as a model plus a harness. The harness, holding prompts,
memory, tools, and control flow, is what a modifier changes; the weights and
the evaluation benchmark usually stay outside the admissible set.
\end{remarkbox}

\vspace{-0.2cm}\paragraph{\protect\circlednumblack{3}~Co-Evolving Evaluators}
The Red Queen G\"odel Machine \citep{iacob2026redqueen} co-evolves the agent
with its evaluator, so the standard that self-improvement is measured against
is itself updated during the process. This places it at the Goal Drift row: the
external objective may remain fixed in name while the mechanism that measures
progress toward it changes. At the far end, fully self-referential
self-improvement, in which no external signal reaches the loop, has been
proposed as a position rather than demonstrated \citep{socratic2024}.

\vspace{-0.2cm}\paragraph{What the GAI Framework Adds}
The nearest prior formal work fixes one design rather than comparing designs:
the G\"odel machine is anchored and proof-gated by construction, and
statistical variants add a risk budget to that gate \citep{sgm2025}. The
recent ladder and the dials answer different questions: the ladder
grades how much of the improvement responsibility the AI carries, and where
the standard sits is not part of the grade; that is what the second dial
reads. Nor does the ladder ask whether the improver is inside the agent; the
first dial does, and it separates systems the ladder groups together.
The defects below follow from the dials rather than
from a rung, and the ladder's own caution that a higher level does not by
itself yield a better improvement process is the same observation read from
the other side \citep{duan2026lastai}. The two dials turn these
choices into coordinates, the polarity classifies the resulting systems, and
the defect catalogue, which we take up next, records what each departure from
the GPI setting costs.

\begin{takeawaybox}
{\normalsize\textbf{Takeaways from existing RSI-labeled work.}}\par
\footnotesize
\begin{itemize}[nosep,leftmargin=1.2em,topsep=2pt]
  \item \textbf{Harness/code is the main substrate.} Most RSI-labeled systems edit prompts, memory, tools, or code, with weights frozen and evaluation outside \citep{yin2024goedelagent,sica2025}.
  \item \textbf{Weight-level work is a second, compute-heavy cluster.} It updates parameters, but usually under a fixed schedule or evaluator; it is not a separate kind of RSI \citep{neohorse2026,lang2026embodirsi}.
  \item \textbf{World-model and rulebook updates are emerging.} A smaller group improves an external world model or rulebook with frozen weights \citep{zhao2026memento3,cui2026rsianything}; Dream-RSI is a steering-policy boundary case \citep{zheng2026dreamrsi}.
  \item \textbf{Self-proposed tasks and data are components, not a paradigm.} They sit inside the modifier; the scoring standard, not the act of self-proposing, decides the polarity, and evaluation remains anchored \citep{tan2026metarsi,meng2026rsigym}.
  \item \textbf{The label does not determine the class.} Here, strict RSI is still the larger group, but several RSI-labeled systems are fixed outer loops under GAI \citep{liu2026solpi,yang2026terminalrsi}.
\end{itemize}
\end{takeawaybox}

\section{Defects of Recursive Self-Improvement}
\label{sec:defects}

The two dials of GAI also locate the failures. The classical guarantees rest on two
conditions: the improvement step is cheap and monotone, and both the
objective and the instrument that evaluates it lie outside the agent. Under
them, the anchored instance converges to an optimal policy in a finite MDP.
Recursive self-improvement may give up both. The defects below are those of the
self-modifying case, $m\in\agent$; the fixed outer loops of \Cref{sec:connections},
including RSI-labeled ones, are outside the scope of this catalogue. Each entry states the condition it violates, is marked
\emph{structural} (it follows from the definitions), and is marked
\emph{observed} where existing systems exhibit it. The list is short by design:
it records where the classical guarantees end, not how far RSI can
go.

\subsection{Search over Candidate Selves and Self-Evaluation}
\label{sec:def-operations}

GPI improves because its improvement step is cheap and its evaluation is
external. The first property is lost when the modifier is an agent component,
the second when the modification critic is. The improvement step is the argmax in \Cref{eq:selfconsistent},
taken over an unbounded content space, where no general procedure decides which
candidate is best. The Gödel machine's proof gate is the established workaround,
admitting only rewrites whose benefit can be proved
\citep{schmidhuber2007goedel}, with statistical variants placing a risk budget
on that same gate \citep{sgm2025}. Evaluation is lost for a related reason: when
$U$ is an agent component, the object evaluated and the instrument of
evaluation coincide, and the value it reports is itself modifiable. A system can
still restrict itself, as the proof gate does, but the restriction is then part
of its content, not an external constraint. Both defects are structural, and
both are also observed. A recent anchored-RSI benchmark where agents revise
their own training-data strategies under a fixed target model and a fixed
external evaluation reports that 58.33\% of settings improve on the first valid
attempt, yet 78.26\% of searches that continue past their best score end with a
lower-scoring final attempt \citep{meng2026rsibench}: the improvement step is
not monotone even when the standard outside is fixed. The second defect is
observed in MASS, which runs its workflow search with the same model as
optimizer and evaluator, so $U\in\agent$, with external judges withheld from
the loop, and reports that the search selects the best self-judged workflow
rather than a certified optimum \citep{lee2026mass}; a world-model RSI system
keeps a revision only after its own component and compatibility checks pass
\citep{cui2026rsianything}.

\subsection{The Consequences of an Ungrounded Base}
\label{sec:def-ungrounded}

When $\rho_U$ is not grounded, the self-consistency conditions constrain only
the pair $(m,U)$: the modifier must propose changes that its own critic rates
highly, and the critic must be tied to its evaluation base. Nothing in the
conditions refers to the goal $G$. Two consequences follow. First, they do not
single out a unique system, and they admit members that serve $G$ poorly or not
at all. Second, they give no faithfulness guarantee: a self-consistent system
may report improvement while the standard it reports against moves, since
calibrating $U$ and honoring $\rho_U$ are choices of content, not properties
enforced from outside.
An engineering audit reaches the same conclusion: from inside
a loop, better capability and a better definition of success give the same
number, so trust is measured by what the system cannot rewrite \citep{tan2026metarsi}.
The case has an early formal analog: Ring and Orseau let an agent rewrite its
own inputs and find that a reinforcement-learning agent can then satisfy its
criterion without the cooperation of the world, the case they call the delusion
box \citep{ring2011delusion}. Both consequences follow from the definitions.

\begin{exbox}
\textbf{Minimal Example (three placements of the same standard).} Consider a
coding agent whose modification critic scores a proposed change by the fraction
of tests it passes, and whose modifier proposes changes that raise that
fraction. The same construction satisfies the self-consistency conditions under
three placements of the tests, which play the role of the base $\rho_U$. Held
out and fixed, they make the reported improvement track the goal of producing
working code, the anchored case. Placed in the agent's own repository, they
admit an improvement that edits the tests, the Goal Drift case of the next
subsection. Replaced by the agent's own judgment of what counts as an
improvement, they depend on the critic itself, the fully self-referential case.
The conditions hold in all three; only the first serves the goal.
\end{exbox}

\subsection{Goal Drift}
\label{sec:def-drift}

The middle row of \Cref{tab:polarity} is the case in which the base still
measures the world and the goal, but the agent may rewrite it. Formally, $G$
stays fixed while $\rho_U$ is an agent component: what moves is the mechanism
that measures progress, not the objective. It is the harder of the two
unanchored settings to notice: the objective keeps its name, and a system that
reports progress against it need not track it. The Red Queen Gödel Machine is an
instance: its evaluator co-evolves with the agent \citep{iacob2026redqueen}. The row has a formal
counterpart: an agent able to rewrite its own utility function is harmless only
when its value function anticipates the rewrite and evaluates the future with
the utility it currently holds \citep{everitt2016selfmod}, a condition secured
for model-based utilities \citep{hibbard2012model}. Both results make the
fixity of the standard something to be secured rather than a default. The defect is structural, and the table marks it as
observed.

\begin{table}[t]
  \centering
  \footnotesize
  \caption{Defects of recursive self-improvement, the dial setting that
    triggers each one, the condition of GPI that it violates, and the support
    for the claim.}
  \label{tab:defects}
  \vspace{-0.2cm}
  \begin{tabularx}{\textwidth}{@{} >{\raggedright\arraybackslash}p{3.0cm}
                     | >{\raggedright\arraybackslash}p{3.0cm}
                     >{\raggedright\arraybackslash\hsize=1.0\hsize}X
                     | >{\raggedright\arraybackslash}p{1.9cm} @{}}
    \toprule
    \textbf{Defect} & \textbf{Trigger} & \textbf{GPI condition violated} &
    \textbf{Status} \\
    \midrule
    Search over candidate selves
      & Dial \circlednum{1}: $m\in\agent$
      & \textcolor{red}{\faTimes}~Improvement is cheap and monotone
      & Structural; observed \\
    \midrule
    Self-evaluation
      & Dial \circlednum{1}: $U\in\agent$
      & \textcolor{red}{\faTimes}~The evaluator stands outside what it evaluates
      & Structural; observed \\
    \midrule
    Ungrounded base
      & Dial \circlednum{2}: $\rho_U$ not grounded
      & \textcolor{red}{\faTimes}~The objective is evaluated from outside the agent
      & Structural \\
    \midrule
    Goal Drift
      & Dial \circlednum{2}: $\rho_U\in\agent$
      & \textcolor{red}{\faTimes}~The objective is evaluated from outside the agent, lost when the base is rewritten
      & Structural; observed \\
    \bottomrule
  \end{tabularx}
  \vspace{-0.2cm}
\end{table}

\Cref{tab:defects} collects the four defects. Each row is a site where a
guarantee of GPI stops applying, and each is either immediate from the
definitions or exhibited by a cited system. Whether a faithfulness monitor
inside the system could detect its own drift is a further question, which we
leave open.

\section{Discussion, Limitations, and Conclusion}
\label{sec:discussion}

\vspace{-0.2cm}\paragraph{Scope and Limitations}

The framework describes how a system changes, not how to compute the change.
The configuration space of \Cref{sec:vocab} is a reference idealization:
writing that an agent instance is drawn from the modifier states what is
chosen, not that the choice is reachable or affordable. Four limits follow.
First, the definitions carry no reachability or complexity claims, and the
defects above are properties of the formulation; where an existing system
exhibits one, we cite it rather than measure it. Second, we do not model time
scales: the alternation between world actions and agent updates is counted in
steps, not compute. Third, practical guards, such as the proof gate of the Gödel
machine or a coding benchmark, are treated as external mechanisms that restrict
what a modifier may propose; whether a guard should itself be an agent component
is a design choice we leave open. Fourth, the framework describes systems that
run without human intervention, the case we study; a reviewer who accepts,
audits, or rolls back an update is outside the formalism, although in practice a
human still sets the direction and reviews the result \citep{favaro2026whenai}.
One objection is worth answering directly: recursive self-improvement is not
simply an MDP over configurations. An MDP fixes its transition law and its
reward outside the agent, whereas here the next configuration is drawn by a
modifier inside the agent, and the standard of evaluation may be inside it as
well.

\vspace{-0.2cm}\paragraph{From Catalogue to Results}

The defects above are stated as consequences of the definitions, and only some
of them are candidates for theorems. Three targets are concrete. First, the
solution set of the self-consistency conditions is uncharacterized: how many
solutions they admit, and when one of them serves the goal, remains open.
Second, the reduction from recursive self-improvement to GPI is prose rather
than a proposition with explicit hypotheses. Third, whether a monitor inside the
system could decide the faithfulness of its own modification critic is left
open, and it is the one question we would expect to require a self-reference
argument rather than a construction. The remaining defects are structural and
would not change under any of these results.

\vspace{-0.2cm}\paragraph{Conclusion}

In this paper, we show that GPI and recursive self-improvement are two settings of one alternating cycle
defined by the GAI formal framework. They differ in two dials: whether the
improving mechanism lies inside the agent, and whether the standard it is
measured against stays grounded. We use these coordinates to place existing
systems, including RSI-labeled ones, on the same two axes, separate those that
are GPI from those that are RSI in the GAI sense, and make the defects of
recursive self-improvement statable one condition at a time.
We believe that this will be a first step toward exploring a formal
characterization of RSI that makes existing
systems comparable, and provides a principled basis for analyzing and designing
new ones.

\setlength{\bibsep}{0.2ex}
\bibliographystyle{assets/plainnat}
\bibliography{references}

\begin{thebibliography}{49}
\providecommand{\natexlab}[1]{#1}
\providecommand{\url}[1]{\texttt{#1}}
\expandafter\ifx\csname urlstyle\endcsname\relax
  \providecommand{\doi}[1]{doi: #1}\else
  \providecommand{\doi}{doi: \begingroup \urlstyle{rm}\Url}\fi

\bibitem[Chen et~al.(2026)Chen, Wang, and Qu]{chen2026rsisurvey}
Mingguang Chen, Licheng Wang, and Bo~Qu.
\newblock Recursive self-improvement in {AI}: From bounded self-refinement to autonomous research loops, 2026.
\newblock arXiv:2607.07663.

\bibitem[Cui et~al.(2026)Cui, Zeng, Xu, Xiao, Xue, Gao, Ouyang, Heng, Yin, Wu, and Yang]{cui2026rsianything}
Taoyong Cui, Pinxian Zeng, Wanghan Xu, Hao Xiao, Shuhan Xue, Qiang Gao, Wanli Ouyang, Pheng~Ann Heng, Zhenfei Yin, Yingcheng Wu, and Ling Yang.
\newblock {RSI-Anything}: Unlocking recursive self-improvement with world models, 2026.
\newblock arXiv:2610.13912.

\bibitem[Dong et~al.(2026)Dong, Zhou, Lin, Wu, Peng, Wang, Fan, Zhang, and Zhao]{dong2026mosib}
Haoyu Dong, Yuhang Zhou, Zihao Lin, Yifan Wu, Bo~Peng, Mingyi Wang, Xiangjun Fan, Lizhu Zhang, and Zhuokai Zhao.
\newblock Mixture of self-improving branches for agent harness optimization, 2026.
\newblock arXiv:2609.37834.

\bibitem[Duan et~al.(2026)Duan, Liu, Tang, Chen, Zhou, Liu, Xu, Wu, Chen, Zhou, Wang, Yu, Han, Zhu, Zhou, Lu, Zhou, Liu, Fang, Tian, Chen, Li, Zuo, Zhang, Yang, Wang, Qiu, He, Li, Zhou, Liu, Wen, Kang, Zhou, and Wu]{duan2026lastai}
Yi~Duan, Ying Liu, Zirui Tang, Haodong Chen, Jun Zhou, Yumou Liu, Bangrui Xu, Yukai Wu, Sidi Chen, Yuhan Zhou, Haoyu Wang, Xiaoyou Yu, Shaokun Han, Xuzhou Zhu, Le~Zhou, Bolin Lu, Wei Zhou, Jiachen Liu, Nuozhou Fang, Jiaxin Tian, Ruoyu Chen, Yuxuan Li, Kai Zuo, Kaiyan Zhang, Qianyu Yang, Zijie Wang, Jiantao Qiu, Conghui He, Guoliang Li, Bowen Zhou, Zhiyuan Liu, Zhoufutu Wen, Jihua Kang, Xuanhe Zhou, and Fan Wu.
\newblock The last {AI} built by humans: Toward genuine recursive self-improvement, 2026.
\newblock arXiv:2609.11873.

\bibitem[Everitt et~al.(2016)Everitt, Filan, Daswani, and Hutter]{everitt2016selfmod}
Tom Everitt, Daniel Filan, Mayank Daswani, and Marcus Hutter.
\newblock Self-modification of policy and utility function in rational agents.
\newblock In \emph{Artificial General Intelligence (AGI)}, volume 9782 of \emph{Lecture Notes in Computer Science}, pages 1--11. Springer, 2016.
\newblock arXiv:1605.03142.

\bibitem[Favaro and Clark(2026)]{favaro2026whenai}
Marina Favaro and Jack Clark.
\newblock When {AI} builds itself, 2026.
\newblock Anthropic Institute, \url{https://www.anthropic.com/institute/recursive-self-improvement}.

\bibitem[Gao et~al.(2025)Gao, Geng, Hua, Hu, Juan, Liu, Liu, Qiu, Qi, Wu, Wang, Xiao, Zhou, Zhang, Zhang, Xiang, Fang, Zhao, Liu, Ren, Qian, Wang, Hu, Wang, Wu, Ji, and Wang]{gao2025survey}
Huan-ang Gao, Jiayi Geng, Wenyue Hua, Mengkang Hu, Xinzhe Juan, Hongzhang Liu, Shilong Liu, Jiahao Qiu, Xuan Qi, Yiran Wu, Hongru Wang, Han Xiao, Yuhang Zhou, Shaokun Zhang, Jiayi Zhang, Jinyu Xiang, Yixiong Fang, Qiwen Zhao, Dongrui Liu, Qihan Ren, Cheng Qian, Zhenhailong Wang, Minda Hu, Huazheng Wang, Qingyun Wu, Heng Ji, and Mengdi Wang.
\newblock A survey of self-evolving agents: What, when, how, and where to evolve on the path to artificial super intelligence, 2025.
\newblock arXiv:2507.21046.

\bibitem[Good(1965)]{good1965}
Irving~John Good.
\newblock Speculations concerning the first ultraintelligent machine.
\newblock In \emph{Advances in Computers}, volume~6, pages 31--88. Academic Press, 1965.

\bibitem[Hibbard(2012)]{hibbard2012model}
Bill Hibbard.
\newblock Model-based utility functions.
\newblock \emph{Journal of Artificial General Intelligence}, 3\penalty0 (1), 2012.
\newblock arXiv:1111.3934.

\bibitem[Hu et~al.(2026)Hu, Hu, Hong, Chen, and Liu]{hu2026coap}
Kairui Hu, Siyuan Hu, Fangzhou Hong, Zhaoxi Chen, and Ziwei Liu.
\newblock Embodied {Turing} machines: Stateful code for robot recursive self-improvement, 2026.
\newblock arXiv:2610.12369.

\bibitem[Iacob et~al.(2026)Iacob, Jovanovi{\'c}, Shen, Burkhardt, Kurmanji, Tastan, Sani, Venanzi, Odonnat, Cao, Marino, Qiu, and Lane]{iacob2026redqueen}
Alex Iacob, Andrej Jovanovi{\'c}, William~F. Shen, Daniel Burkhardt, Meghdad Kurmanji, Nurbek Tastan, Lorenzo Sani, Niccol{\`o} Alberto~Elia Venanzi, Ambroise Odonnat, Zeyu Cao, Bill Marino, Xinchi Qiu, and Nicholas~D. Lane.
\newblock The {Red Queen} {G\"o}del machine: Co-evolving agents and their evaluators, 2026.
\newblock arXiv:2606.26294.

\bibitem[Kakade et~al.(2026)Kakade, Srivastava, and Karande]{kakade2026polaris}
Aditya Kakade, Vivek Srivastava, and Shirish Karande.
\newblock Polaris: A {G\"o}del agent framework for small language models through experience-abstracted policy repair, 2026.
\newblock arXiv:2603.23129.

\bibitem[Lang et~al.(2026)Lang, Lu, Sang, Luo, Chen, Li, Yu, Shi, and Wang]{lang2026embodirsi}
Haoran Lang, Haotao Lu, Shiyu Sang, Haoyang Luo, Guo Chen, Qun Li, Jingyi Yu, Ye~Shi, and Jingya Wang.
\newblock {EmbodiRSI}: Recursive self-improvement for data-efficient robot adaptation, 2026.
\newblock arXiv:2609.38905.

\bibitem[Lee et~al.(2026)Lee, Xu, Seely, Lee, Sojoudi, Zaharia, and Tang]{lee2026mass}
Hyunin Lee, Jinglue Xu, Jeffrey Seely, Donghyun Lee, Somayeh Sojoudi, Matei Zaharia, and Yujin Tang.
\newblock Recursive self-improvement through multi-agent self-supervision, 2026.
\newblock arXiv:2610.12176.

\bibitem[Liu et~al.(2026)Liu, Ye, Gao, Cao, Li, Zhuge, Wang, Zhang, Luo, Bian, Zhu, Zhu, Xie, and Han]{liu2026solpi}
Haozhe Liu, Tian Ye, Sensen Gao, Qihang Cao, Yitong Li, Mingchen Zhuge, Duomin Wang, Ruihua Zhang, Ping Luo, Jiawang Bian, Lei Zhu, Ligeng Zhu, Enze Xie, and Song Han.
\newblock {SoL-Pi}: Recursively scaling auto-research loops for efficient agent harness, 2026.
\newblock arXiv:2609.20519.

\bibitem[Lu et~al.(2024)Lu, Lu, Lange, Foerster, Clune, and Ha]{lu2024aiscientist}
Chris Lu, Cong Lu, Robert~Tjarko Lange, Jakob Foerster, Jeff Clune, and David Ha.
\newblock The {AI} scientist: Towards fully automated open-ended scientific discovery, 2024.
\newblock arXiv:2408.06292.

\bibitem[Madaan et~al.(2023)Madaan, Tandon, Gupta, Hallinan, Gao, Wiegreffe, Alon, Dziri, Prabhumoye, Yang, Gupta, Majumder, Hermann, Welleck, Yazdanbakhsh, and Clark]{madaan2023selfrefine}
Aman Madaan, Niket Tandon, Prakhar Gupta, Skyler Hallinan, Luyu Gao, Sarah Wiegreffe, Uri Alon, Nouha Dziri, Shrimai Prabhumoye, Yiming Yang, Shashank Gupta, Bodhisattwa~Prasad Majumder, Katherine Hermann, Sean Welleck, Amir Yazdanbakhsh, and Peter Clark.
\newblock Self-{R}efine: Iterative refinement with self-feedback.
\newblock In \emph{Advances in Neural Information Processing Systems}, 2023.
\newblock arXiv:2303.17651.

\bibitem[Meng et~al.(2026{\natexlab{a}})Meng, Du, Chen, Zhao, Lu, Hu, and Shieh]{meng2026rsibench}
Fanqing Meng, Lingxiao Du, Qiguang Chen, Ziqi Zhao, Haocheng Lu, Mengkang Hu, and Michael~Qizhe Shieh.
\newblock {RSIBench-Data}: Benchmarking data-centric research for recursive self-improvement, 2026{\natexlab{a}}.
\newblock arXiv:2607.25886.

\bibitem[Meng et~al.(2026{\natexlab{b}})Meng, Du, Lu, Chen, Zhao, Wu, Zhuo, Hu, and Shieh]{meng2026rsigym}
Fanqing Meng, Lingxiao Du, Haocheng Lu, Qiguang Chen, Ziqi Zhao, Zijian Wu, Jiayuan Zhuo, Mengkang Hu, and Michael~Qizhe Shieh.
\newblock {RSIGym}: A flexible environment for recursive self-improvement, 2026{\natexlab{b}}.
\newblock arXiv:2610.10310.

\bibitem[Mnih et~al.(2013)Mnih, Kavukcuoglu, Silver, Graves, Antonoglou, Wierstra, and Riedmiller]{mnih2013dqn}
Volodymyr Mnih, Koray Kavukcuoglu, David Silver, Alex Graves, Ioannis Antonoglou, Daan Wierstra, and Martin Riedmiller.
\newblock Playing {A}tari with deep reinforcement learning.
\newblock In \emph{NIPS Deep Learning Workshop}, 2013.
\newblock arXiv:1312.5602.

\bibitem[{NeoHorse Team} et~al.(2026){NeoHorse Team}, Cao, Dai, Guo, Han, Hu, Jiang, Kuang, Li, Li, Pei, Tian, Wang, Wang, Wang, Wu, Xu, Zhang, Zhou, Cheng, Fan, He, Jiao, Li, Li, Liu, Liu, Liu, Pan, Ren, Song, Wang, Yu, Zhang, Zhang, Zheng, and Zong]{neohorse2026}
{NeoHorse Team}, Guoliang Cao, Guohao Dai, Tianyu Guo, Kai Han, Hailin Hu, Zihan Jiang, Xiang Kuang, Boxun Li, Yulong Li, Zehua Pei, Yuchuan Tian, Jiamin Wang, Yu~Wang, Yunhe Wang, Yihong Wu, Haiyang Xu, Shuo Zhang, Hang Zhou, Siyang Cheng, Jiayu Fan, Wei He, Qingrui Jiao, Hongguang Li, Zhiyuan Li, Runke Liu, Xi~Liu, Xinchen Liu, Sinno~Jialin Pan, Yi~Ren, Liuyang Song, Chenyu Wang, Bei Yu, Quanlu Zhang, Xiangyu Zhang, Mengyu Zheng, and Yingjie Zong.
\newblock {NeoHorse-1}: Towards recursive self-improvement via agentic post-training with routing harness, 2026.
\newblock arXiv:2609.08183.

\bibitem[Nivel et~al.(2013)Nivel, Th{\'o}risson, Steunebrink, Dindo, Pezzulo, Rodriguez, Hernandez, Ognibene, Schmidhuber, Sanz, Helgason, Chella, and Jonsson]{nivel2013bounded}
E.~Nivel, K.~R. Th{\'o}risson, B.~R. Steunebrink, H.~Dindo, G.~Pezzulo, M.~Rodriguez, C.~Hernandez, D.~Ognibene, J.~Schmidhuber, R.~Sanz, H.~P. Helgason, A.~Chella, and G.~K. Jonsson.
\newblock Bounded recursive self-improvement, 2013.
\newblock arXiv:1312.6764.

\bibitem[Novikov et~al.(2025)Novikov, V{\~u}, Eisenberger, Dupont, Huang, Wagner, Shirobokov, Kozlovskii, Ruiz, Mehrabian, Kumar, See, Chaudhuri, Holland, Davies, Nowozin, Kohli, and Balog]{novikov2025alphaevolve}
Alexander Novikov, Ng{\^a}n V{\~u}, Marvin Eisenberger, Emilien Dupont, Po-Sen Huang, Adam~Zsolt Wagner, Sergey Shirobokov, Borislav Kozlovskii, Francisco J.~R. Ruiz, Abbas Mehrabian, M.~Pawan Kumar, Abigail See, Swarat Chaudhuri, George Holland, Alex Davies, Sebastian Nowozin, Pushmeet Kohli, and Matej Balog.
\newblock {AlphaEvolve}: A coding agent for scientific and algorithmic discovery, 2025.
\newblock arXiv:2506.13131.

\bibitem[Omohundro(2008)]{omohundro2008basic}
Stephen~M. Omohundro.
\newblock The basic {AI} drives.
\newblock In \emph{Artificial General Intelligence}, volume 171 of \emph{Frontiers in Artificial Intelligence and Applications}, pages 483--492. IOS Press, 2008.

\bibitem[Ring and Orseau(2011)]{ring2011delusion}
Mark Ring and Laurent Orseau.
\newblock Delusion, survival, and intelligent agents.
\newblock In \emph{Artificial General Intelligence (AGI)}, volume 6830 of \emph{Lecture Notes in Computer Science}, pages 11--20. Springer, 2011.

\bibitem[Robeyns et~al.(2025)Robeyns, Szummer, and Aitchison]{sica2025}
Maxime Robeyns, Martin Szummer, and Laurence Aitchison.
\newblock A self-improving coding agent, 2025.
\newblock arXiv:2504.15228.

\bibitem[Schaul(2024)]{socratic2024}
Tom Schaul.
\newblock Boundless {S}ocratic learning with language games, 2024.
\newblock arXiv:2411.16905.

\bibitem[Schmidhuber(2003)]{schmidhuber2003goedeltech}
J{\"u}rgen Schmidhuber.
\newblock {G\"o}del machines: Self-referential universal problem solvers making provably optimal self-improvements, 2003.
\newblock arXiv:cs/0309048.

\bibitem[Schmidhuber(2007)]{schmidhuber2007goedel}
J{\"u}rgen Schmidhuber.
\newblock G{\"o}del machines: Fully self-referential optimal universal self-improvers.
\newblock In Ben Goertzel and Cassio Pennachin, editors, \emph{Artificial General Intelligence}, pages 199--226. Springer, 2007.

\bibitem[Schulman et~al.(2017)Schulman, Wolski, Dhariwal, Radford, and Klimov]{schulman2017ppo}
John Schulman, Filip Wolski, Prafulla Dhariwal, Alec Radford, and Oleg Klimov.
\newblock Proximal policy optimization algorithms, 2017.
\newblock arXiv:1707.06347.

\bibitem[Shinn et~al.(2023)Shinn, Cassano, Berman, Gopinath, Narasimhan, and Yao]{shinn2023reflexion}
Noah Shinn, Federico Cassano, Edward Berman, Ashwin Gopinath, Karthik Narasimhan, and Shunyu Yao.
\newblock Reflexion: Language agents with verbal reinforcement learning.
\newblock In \emph{Advances in Neural Information Processing Systems}, 2023.
\newblock arXiv:2303.11366.

\bibitem[Srikanth et~al.(2026)Srikanth, Zhao, Xu, Wu, and Jiang]{srikanth2026aide2}
Dhruv Srikanth, Bingchen Zhao, Dixing Xu, Yuxiang Wu, and Zhengyao Jiang.
\newblock Recursive self-improvement of {AI} research agents, 2026.
\newblock arXiv:2609.26457.

\bibitem[Sutton and Barto(2018)]{sutton2018rl}
Richard~S. Sutton and Andrew~G. Barto.
\newblock \emph{Reinforcement Learning: An Introduction}.
\newblock MIT Press, 2nd edition, 2018.

\bibitem[Tan et~al.(2026)Tan, Sun, Shi, Liu, Wu, Brooks, Qian, Shang, Huang, Ding, Liao, Li, Sun, Liu, Rong, Yu, Chen, Geng, Li, Wang, Tu, Tang, Zhan, Jiang, Wu, Zhang, Liang, Wu, Zhang, Marlow, and Wan]{tan2026metarsi}
Zihan Tan, Leixin Sun, Zitong Shi, Yitao Liu, Jiajun Wu, Nathaniel Brooks, Jiaru Qian, Xiaoran Shang, Suyuan Huang, Yi~Ding, Yangxu Liao, Mukai Li, Qiushi Sun, Shudong Liu, Xuankun Rong, Xiaohang Yu, Zhuo Chen, Hejia Geng, Chenxin Li, Aozhou Wang, Zengji Tu, Robert Tang, Yuxin Zhan, Eric Jiang, Yuxin Wu, Jianqing Zhang, Xiao Liang, Fang Wu, Haochi Zhang, Alexander Marlow, and Guancheng Wan.
\newblock {MetaRSI} / {RSI}$^{2}$: A meta-recursive self-improving system for recursive self-improving systems themselves, 2026.
\newblock arXiv:2609.06396.

\bibitem[Tang et~al.(2026)Tang, Wang, Zhang, Wang, Chen, Zhang, Ma, Jiang, Guo, Li, Jiao, Cui, Yin, Liang, Wan, Lin, Zhao, Zhang, Zhou, Zhang, and You]{tang2026trinity}
Weidong Tang, Ruiheng Wang, Jingxuan Zhang, Hexin Wang, Haoyuan Chen, Zhengfeiyang Zhang, Yixing Ma, Zichen Jiang, Ziyu Guo, Haodong Li, Zhengbo Jiao, Zhiqing Cui, Xinxiang Yin, Shuo Liang, Zhenglin Wan, Hongzhan Lin, Wangbo Zhao, Kaipeng Zhang, Pengfei Zhou, Odin Zhang, and Yang You.
\newblock The {Trinity}: Capability-driven recursive self-improvement for multimodal agents, 2026.
\newblock arXiv:2610.36077.

\bibitem[Tian et~al.(2026)Tian, Zhang, Liu, Liu, Liu, and Wang]{tian2026physevo}
Wenqing Tian, Zeyu Zhang, Zhaocheng Liu, Fengwei Liu, Qiang Liu, and Liang Wang.
\newblock {PhysEvo}: Astra can act, let it, 2026.
\newblock arXiv:2610.08995.

\bibitem[Wu et~al.(2025)Wu, Yin, Kang, Zhang, Xu, Chen, and Zhang]{sgm2025}
Xuening Wu, Shenqin Yin, Yanlan Kang, Xinhang Zhang, Qianya Xu, Zeping Chen, and Wenqiang Zhang.
\newblock {SGM}: A statistical {G\"o}del machine for risk-controlled recursive self-modification, 2025.
\newblock arXiv:2510.10232.

\bibitem[Yampolskiy(2015)]{yampolskiy2015seed}
Roman~V. Yampolskiy.
\newblock From seed {AI} to technological singularity via recursively self-improving software, 2015.
\newblock arXiv:1502.06512.

\bibitem[Yang et~al.(2026)Yang, Li, Shi, Li, Wang, Zhou, Mi, Yan, and Liang]{yang2026terminalrsi}
Junyao Yang, Zongxia Li, Yucheng Shi, Zhongzhi Li, Ruhan Wang, Ruiwen Zhou, Haitao Mi, Shuicheng Yan, and Leowei Liang.
\newblock {Terminal-RSI}: Recursive self-improving terminal agents via recursive self-rewrite and policy-refine, 2026.
\newblock arXiv:2610.22192.

\bibitem[Ye et~al.(2026)Ye, Lu, Dong, Su, and Song]{ye2026harnesszero}
Haoran Ye, Yuxing Lu, Haonan Dong, Zhaochen Su, and Guojie Song.
\newblock Harness-zero: Harness distillation via agent-as-harness, 2026.
\newblock arXiv:2609.24974.

\bibitem[Yin et~al.(2024)Yin, Wang, Pan, Lin, Wan, and Wang]{yin2024goedelagent}
Xunjian Yin, Xinyi Wang, Liangming Pan, Li~Lin, Xiaojun Wan, and William~Yang Wang.
\newblock G{\"o}del agent: A self-referential agent framework for recursive self-improvement, 2024.
\newblock arXiv:2410.04444.

\bibitem[Yudkowsky(2013)]{yudkowsky2013micro}
Eliezer Yudkowsky.
\newblock Intelligence explosion microeconomics, 2013.
\newblock MIRI Technical Report.

\bibitem[Zelikman et~al.(2023)Zelikman, Lorch, Mackey, and Kalai]{zelikman2023stop}
Eric Zelikman, Eliana Lorch, Lester Mackey, and Adam~Tauman Kalai.
\newblock Self-taught optimizer ({STOP}): Recursively self-improving code generation, 2023.
\newblock arXiv:2310.02304.

\bibitem[Zhang et~al.(2025)Zhang, Hu, Lu, Lange, and Clune]{zhang2025darwin}
Jenny Zhang, Shengran Hu, Cong Lu, Robert Lange, and Jeff Clune.
\newblock Darwin {G\"o}del machine: Open-ended evolution of self-improving agents, 2025.
\newblock arXiv:2505.22954.

\bibitem[Zhang et~al.(2026)Zhang, Zhao, Yang, Foerster, Clune, Jiang, Devlin, and Shavrina]{zhang2026hyperagents}
Jenny Zhang, Bingchen Zhao, Wannan Yang, Jakob Foerster, Jeff Clune, Minqi Jiang, Sam Devlin, and Tatiana Shavrina.
\newblock Hyperagents, 2026.
\newblock arXiv:2603.19461.

\bibitem[Zhang(2026)]{zhang2026rsiblog}
Yuxuan Zhang.
\newblock {RSI}: A conditional theory, not an achieved milestone, 2026.
\newblock Blog post, \url{https://www.cs.ubc.ca/~reacher/blog/rsi/}.

\bibitem[Zhao et~al.(2026{\natexlab{a}})Zhao, Yu, He, Fang, Tutunov, Bou-Ammar, Luo, and Wang]{zhao2026memento3}
Haoyu Zhao, Zhengxu Yu, Zhiyuan He, Meng Fang, Rasul Tutunov, Haitham Bou-Ammar, Weilin Luo, and Jun Wang.
\newblock {Memento} 3: Model-based recursive self-improvement through reflective rulebooks, 2026{\natexlab{a}}.
\newblock arXiv:2610.11794.

\bibitem[Zhao et~al.(2026{\natexlab{b}})Zhao, Zhou, Qi, Fu, Zheng, Xia, Wang, Xuan, and Yokoya]{zhao2026rsi4eo}
Sijie Zhao, Zeqi Zhou, Heli Qi, Xuanchao Fu, Zhuo Zheng, Junshi Xia, Junjue Wang, Weihao Xuan, and Naoto Yokoya.
\newblock The last remote sensing model built by humans? casting any earth observation task as recursive self-improvement, 2026{\natexlab{b}}.
\newblock arXiv:2610.18845.

\bibitem[Zheng et~al.(2026)Zheng, Wu, Zhang, He, Zhang, Coleman, Wei, Bai, Liu, Liu, Wang, Zhuan, Kang, Xiang, Huang, Cheng, and Guo]{zheng2026dreamrsi}
Tong Zheng, Xidong Wu, Zheng Zhang, Zhankui He, Chaoyi Zhang, Benjamin Coleman, Ruoqiao Wei, Di~Bai, Haolin Liu, Rui Liu, Xue Wang, Yue Zhuan, Wang-Cheng Kang, Renkai Xiang, Heng Huang, Xinwu Cheng, and Yunsong Guo.
\newblock {Dream-RSI}: Recursive self-improvement through evolving worlds, 2026.
\newblock arXiv:2609.14858.

\end{thebibliography}

\clearpage
\appendix

\begin{center}
    {\Huge\bfseries\textcolor{appendixpurple}{Appendix: Notation}}
\end{center}
\vspace{0.6em}
\phantomsection\label{app:notation}

\begin{tcolorbox}[enhanced, breakable, colback=appendixlightpurple,
    colframe=appendixpurple, boxrule=0.8pt, arc=2pt,
    left=1.1em, right=1.1em, top=1.0em, bottom=1.0em]
\small
\renewcommand{\arraystretch}{1.2}
\textbf{World, goal, and RL background.}
\begin{center}
\begin{tabularx}{\textwidth}{@{}llX@{}}
\toprule
\textbf{Symbol} & & \textbf{Meaning} \\
\midrule
$\cW$ & & world $=(\cS,\cA,p,r,\gamma)$; states, actions, transition, reward, discount \\
$\cS,\cA$ & & world state / action spaces \\
$r$ & & world reward; the grounded content of $\rho_V$ (\Cref{sec:vocab}) \\
$G$ & & external goal / preference; the grounded content of $\rho_U$ \\
$\gamma$ & & discount factor \\
$\mathbb{E}$ & & expectation \\
$\cE_m,\cI_m$ & & evaluation / improvement operators carried by the modifier \\
\bottomrule
\end{tabularx}
\end{center}

\vspace{0.9em}
\textbf{Components and roles.}
\begin{center}
\begin{tabularx}{\textwidth}{@{}llX@{}}
\toprule
\textbf{Symbol} & & \textbf{Meaning} \\
\midrule
$\cO$ & & set of components in the system \\
$o$ & & a component; $\cC_o$ its content space \\
$\pi$ & & policy: chooses world actions, $\cS\to\Dist{\cA}$ \\
$V$ & & action critic: evaluation of the policy, $\cS\to\R$ \\
$m$ & & modifier: proposes how the system changes, $\cfgspace\to\Dist{\agentspace}$ \\
$U$ & & modification critic: evaluation of the modifier, $\cfgspace\to\R$ \\
$\rho$ & & evaluation base of the critics; instances $\rho_V,\rho_U$ where the two must be told apart \\
\bottomrule
\end{tabularx}
\end{center}

\vspace{0.9em}
\textbf{System and agent.}
\begin{center}
\begin{tabularx}{\textwidth}{@{}llX@{}}
\toprule
\textbf{Symbol} & & \textbf{Meaning} \\
\midrule
$\cfg$ & & a system, $\cfg\in\cfgspace$ \\
$\cfgspace$ & & full system space, $\cfgspace=\prod_{o\in\cO}\cC_o$ \\
$\agent\subseteq\cO$ & & the agent: the set of modifiable components \\
$\agentspace$ & & agent subspace, $\agentspace=\prod_{o\in\agent}\cC_o\subseteq\cfgspace$ \\
$\agentst$ & & an agent instance, $\agentst\in\agentspace$ \\
$\replace{\agentst}$ & & system with its agent part replaced by $\agentst$ \\
$\Dist{Y}$ & & set of probability distributions over $Y$ \\
\bottomrule
\end{tabularx}
\end{center}
\end{tcolorbox}

\end{document}